\documentclass[letterpaper, 10 pt, conference]{ieeeconf}  

\IEEEoverridecommandlockouts                              

\usepackage{graphics} 
\usepackage{epsfig} 
\usepackage{times} 
\usepackage{amsmath} 
\usepackage{amssymb}  

\usepackage{graphicx}
\usepackage{tabularx}
\usepackage{caption} 
\usepackage{wrapfig}
\usepackage{romannum}
\usepackage{bm}
\usepackage[table,xcdraw]{xcolor}
\usepackage{tabularray}

\usepackage{multirow}
\usepackage{booktabs}
\usepackage{pifont}
\usepackage{bbm} 

\usepackage{xcolor}

\newcommand{\bpi}{\mathbf{p}_i}
\newcommand{\bbi}{\mathbf{b}_i}
\newcommand{\bbstar}{\mathbf{b}_*}
\newcommand{\byi}{\mathbf{y}_i}

\definecolor{puddle}{rgb}{0.0, 0.0, 1.0}
\definecolor{object}{rgb}{0.8, 0.6, 1.0}
\definecolor{paved}{rgb}{1.0, 1.0, 0.0}
\definecolor{unpaved}{rgb}{1.0, 0.6, 0.8}
\definecolor{dirt}{rgb}{0.6, 0.298039, 0.0}
\definecolor{grass}{rgb}{0.435294, 1.0, 0.290196}
\definecolor{vegetation}{rgb}{0.0, 0.4, 0.0}

\newcommand\semcolor[1][black]{\fcolorbox{black}{#1}{\rule{0mm}{1mm}\rule{1mm}{0mm}}}

\newcommand\semcolorbf[2]{\textbf{\textcolor{#1}{#2}}}

\newcommand{\ourprevWork}{EBS}
\newcommand{\ourdataName}{OffRoad}

\newcommand{\ourdatabf}{\textbf{\ourdataName}}

\newcommand{\calL}{\mathcal{L}}

\newcommand{\Dkl}{D_{\text{KL}}}

\newcommand{\datasetSem}{\mathcal{D}_{\text{sem}} = \{v_i, \byi\}^N_{i=1}}

\newcommand{\Uthres}{\mathrm{U}_{\text{thr}}}

\newcommand{\eg}{\textit{e}.\textit{g}.}

\newcommand{\rellisC}{RELLIS-3D~\cite{122_RELLIS_jiang2021rellis} }

\newcommand{\dol}{\frac{d}{l}}

\newcommand{\Tref}[1]{Table~\ref{#1}}
\newcommand{\Fref}[1]{Fig.~\ref{#1}}

\DeclareCaptionFont{mysize}{\fontsize{8}{9.6}\selectfont}
\makeatletter
\let\NAT@parse\undefined
\makeatother
\usepackage[colorlinks=true]{hyperref}

\title{\LARGE \bf
Uncertainty-aware Semantic Mapping in Off-road Environments \\ with Dempster-Shafer Theory of Evidence
}

\author{Junyoung Kim and Junwon Seo%
\thanks{J. Kim and J. Seo are with the Agency for Defense Development (ADD), Daejeon 34186, Republic of Korea. {\tt\footnotesize \{junyoung.kimv, junwon.vision\}@gmail.com}}%
}

\begin{document}

\maketitle

\begin{abstract}
Semantic mapping with Bayesian Kernel Inference~(BKI) has shown promise in providing a richer understanding of environments by effectively leveraging local spatial information. However, existing methods face challenges in constructing accurate semantic maps or reliable uncertainty maps in perceptually challenging environments due to unreliable semantic predictions. To address this issue, we propose an evidential semantic mapping framework, which integrates the evidential reasoning of Dempster-Shafer Theory of Evidence~(DST) into the entire mapping pipeline by adopting Evidential Deep Learning~(EDL) and Dempster's rule of combination. Additionally, the extended belief is devised to incorporate local spatial information based on their uncertainty during the mapping process. Comprehensive experiments across various off-road datasets demonstrate that our framework enhances the reliability of uncertainty maps, consistently outperforming existing methods in scenes with high perceptual uncertainties while showing semantic accuracy comparable to the best-performing semantic mapping techniques.
\end{abstract}
\section{INTRODUCTION}
\newcommand{\citeOccupancyMap}{\cite{131_OccupancyMap_elfes1989using, 6_OctoMap_hornung2013octomap}}
\newcommand{\citeSparsity}{\cite{50_GPOctoMap_wang2016fast}}
\newcommand{\citeSemanticMap}{\cite{94_kim20133d, 103_valentin2013mesh, 95_SemanticOctree_sengupta2015semantic, 97_yang2017semantic, 98_DA-RNN_xiang2017rnn, 96_SemanticFusion_mccormac2017semanticfusion, 109_paz2020probabilistic, 23_SSMI_asgharivaskasi2023semantic, 26_morilla2023robust, 106_FusionOverconfidence_marques2023overconfidence}}
\newcommand{\citeSemanticMapCompact}{\cite{94_kim20133d, 103_valentin2013mesh, 95_SemanticOctree_sengupta2015semantic, 109_paz2020probabilistic, 23_SSMI_asgharivaskasi2023semantic, 26_morilla2023robust, 106_FusionOverconfidence_marques2023overconfidence}}
\newcommand{\citeVoxelSemanticMap}{\cite{94_kim20133d, 95_SemanticOctree_sengupta2015semantic, 97_yang2017semantic, 98_DA-RNN_xiang2017rnn, 23_SSMI_asgharivaskasi2023semantic, 26_morilla2023robust, 106_FusionOverconfidence_marques2023overconfidence}}
\newcommand{\citeDNNbasedSemanticMap}{\cite{96_SemanticFusion_mccormac2017semanticfusion, 98_DA-RNN_xiang2017rnn, 99_RecurrentOctoMap_sun2018recurrent, 105_maturana2018real, wilson2022motionsc}}
\newcommand{\citeEDLVariousField}{\cite{70_USNet_chang2022fast, 72_TMCJournal_han2022trusted, 13_EvPSNet_sirohi2023uncertainty, 120_EVORA_cai2023evora}}

Semantic mapping aims to construct detailed representations of the environment by estimating voxel-wise semantic states~\citeSemanticMapCompact. To alleviate the issue of discontinuous maps when utilizing sparse sensors such as LiDAR~\citeSparsity, techniques that utilize local spatial information have been proposed. Especially, the adoption of Bayesian Kernel Inference~(BKI)~\cite{52_BKI_vega2014nonparametric} has enabled the efficient utilization of neighboring measurements, resulting in more continuous semantic maps~\cite{51_S-BKI_gan2020bayesian, unnikrishnan2022dynamic, wilson2023convolutional}.

Recent approaches extract semantic information from sensor measurements by utilizing Deep Neural Networks~(DNNs)~\citeDNNbasedSemanticMap, which often yield unreliable predictions in off-road scenes~\cite{86_OFFROAD_jin2021memory, seo2023learning, kim2024evidential, frey2024roadrunner}. Thus, it is essential to assess the dependability of semantic predictions during the mapping process for enhanced reliability in perceptually challenging off-road environments. Though the dependability of predictions can be evaluated with their uncertainty, conventional DNNs are known to be overconfident~\cite{107_guo2017calibration, 83_ModernReliability_de2023reliability}, necessitating techniques for uncertainty quantification. Among various techniques, Evidential Deep Learning~(EDL)~\cite{10_EDL_sensoy2018evidential} provides an efficient method to estimate the uncertainty of DNN predictions.

Based on the predictive uncertainty, the mapping framework can adaptively prioritize confident predictions to enhance the reliability of information and reduce inaccuracies~\cite{26_morilla2023robust}. This also allows for a more accurate estimation of map cell uncertainty, leading to safer deployments for downstream tasks. Following this concept, our prior work introduced an uncertainty-aware BKI framework in~\cite{kim2024evidential}, showing enhanced semantic mapping accuracy. Nonetheless, the BKI-based mapping framework estimates map cell uncertainty based on the variance of Dirichlet posterior without directly incorporating semantic uncertainty into its calculations~\cite{51_S-BKI_gan2020bayesian, 47_ConvBKI2_wilson2023convbki}. Therefore, the map cell uncertainty derived from the BKI-based mapping primarily focuses on geometric uncertainty, failing to reflect semantic uncertainty adequately.

The reliable quantification of map cell uncertainty is crucial for safe-critical applications~\cite{Fan-RSS-21, 47_ConvBKI2_wilson2023convbki}. Since the number of semantic classes is limited, the robots can plan their trajectory within the same semantic region based on the uncertainty information to maximize their success rate. Moreover, uncertainty information can guide active exploration or object-goal navigation~\cite{106_FusionOverconfidence_marques2023overconfidence} by distinguishing regions with low and high uncertainty. This highlights the importance of developing approaches that integrate semantic uncertainty more effectively into the mapping process.

\begin{figure*}[t]
\begin{center}
\includegraphics[width=1.0\linewidth]{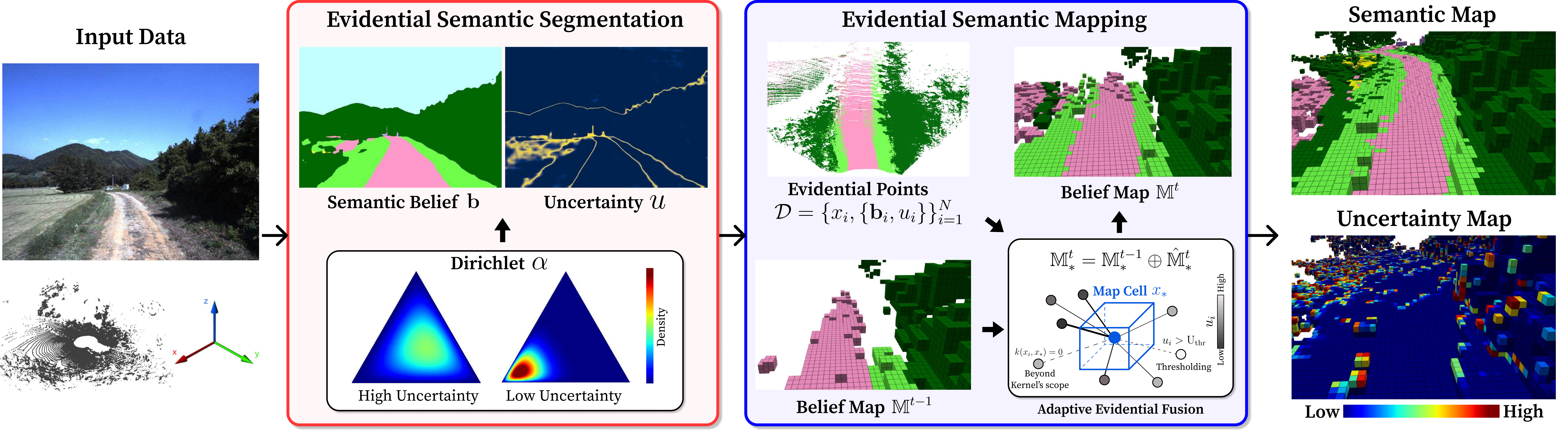}
\end{center}
\vspace{-0.1in}
\caption{Overview of our evidential semantic mapping framework. With an evidential segmentation network trained by EDL~\cite{10_EDL_sensoy2018evidential}, input data is processed to derive continuous semantic belief and uncertainty. These 3D evidential points are then integrated into the semantic map through adaptive evidential fusion using Dempster's rule of combination, resulting in a dependable semantic map and uncertainty map in environments with high perceptual uncertainties.}
\label{fig:framework}
\vspace{-0.1in}
\end{figure*}

In this work, we propose an evidential semantic mapping framework for reliable deployments in perceptually challenging off-road environments. We devise our framework to fully integrate semantic uncertainty estimated via EDL into the mapping process by leveraging Dempster-Shafer Theory of Evidence~(DST)~\cite{rakowsky2007fundamentals}, as shown in \Fref{fig:framework}. This includes adopting Dempster's rule of combination to ensure the entire framework is grounded in DST, enabling systematic information integration based on uncertainty. Consequently, our framework accurately reflects semantic uncertainty in the uncertainty map compared to BKI-based mapping. We evaluate our framework against existing continuous semantic mapping methods using \rellisC and our off-road datasets and show that our framework estimates map cell uncertainty more reliably than the BKI-based methods while performing comparably regarding semantic mapping accuracy.

\section{RELATED WORK}
\subsection{Semantic Mapping}
As BKI~\cite{52_BKI_vega2014nonparametric} enables efficient integration of local spatial information into the mapping process~\cite{49_BGKOctoMap_doherty2017bayesian}, S-BKI~\cite{51_S-BKI_gan2020bayesian} introduces the BKI-based continuous mapping into the semantic mapping field. While various approaches have been proposed to improve the accuracy of continuous semantic mapping~\cite{47_ConvBKI2_wilson2023convbki,88_SEE-CSOM_deng2023see}, the uncertainty of semantic predictions has not been incorporated into the mapping, leading to unreliable mapping in challenging environments such as off-road conditions. Although our previous work~\cite{kim2024evidential} proposes an uncertainty-aware BKI extension to incorporate semantic uncertainty for robust deployments in off-road environments with high perceptual uncertainties, the map cell uncertainty in BKI does not utilize semantic uncertainty directly as it is defined as the variance of Dirichlet posterior.

\subsection{Evidential Deep Learning}
Leveraging DST~\cite{129_DST_yager2008classic}, the EDL framework models a second-order Dirichlet distribution from which the predictive uncertainty is computed~\cite{10_EDL_sensoy2018evidential, 54_IEDL_deng2023uncertainty, ulmer2023prior}. Since DST also allows the outputs of EDL to be integrated based on their uncertainty via Dempster's combination rule, the EDL framework can be utilized in tasks that require the accumulation of information from different sources, such as multiple time-steps, sensors, or views~\cite{72_TMCJournal_han2022trusted, han2023ds}. Especially, the rule effectively fuses the coherent parts and ignores the conflicting information, enabling accurate reasoning in DST. Consequently, EDL has been adopted in various fields for uncertainty estimation~\citeEDLVariousField. In this work, we adopt and refine Dempster's rule of combination~\cite{72_TMCJournal_han2022trusted} to effectively utilize semantic uncertainty and local spatial information during the mapping process.

\section{METHODS}
\subsection{Evidential Semantic Segmentation}\label{METHOD_EDL}
To utilize the semantic uncertainty of predictions during the mapping process, we apply the EDL framework~\cite{10_EDL_sensoy2018evidential} to existing segmentation networks to reliably estimate its uncertainty. The output of the EDL network for $i$-th input is regarded as an evidence vector $\mathbf e_i = [e_i^1, ..., e_i^C]$, where $C$ is the number of semantic classes. From the evidence vector $\mathbf e_i$, belief mass $b_i^c = e_i^c / S_i$ and its overall uncertainty $u_i = C / S_i$ are obtained by Subjective Logic~\cite{130_SubjectiveLogic_jsang2018subjective} with the following property:
\vspace{-0.1in}
\begin{align}\label{belief_sum}
    u_i + \sum^C_{c=1} b_i^c = 1,    
\end{align}
where $S_i = \sum^C_{c=1}(e_i^c+1)$. Then, the semantic probability $\bpi = [p_i^1, ..., p_i^C]$ can be modeled as $p_i^c = \alpha_i^c / S_i$ by Dirichlet distribution with concentration parameters $\boldsymbol{\alpha}_i = [\alpha_i^1, ..., \alpha_i^C]$, where $\alpha_i^c = e_i^c + 1$.

We adopt the loss from~\cite{10_EDL_sensoy2018evidential, 13_EvPSNet_sirohi2023uncertainty} to train the evidential segmentation network with EDL. The following evidential loss encourages collecting semantic evidence of its corresponding one-hot label $\byi$ from the training dataset $\datasetSem$ for each pixel $v_i$:
\begin{equation}\label{EDL_evidential_loss}
\begin{aligned}
    \calL_{\text{EDL}}^i = \sum^C_{c=1} y_i^c (\log(S_i) - \log(\alpha_i^c)).
\end{aligned}
\end{equation}
Additionally, KL regularization term~\cite{10_EDL_sensoy2018evidential} is utilized to penalize incorrect evidence on non-ground-truth class as follows:
\begin{align}
    \calL_{\text{Reg}}^i =  \Dkl\bigl[\text{Dir}\left(\bpi | \boldsymbol{\tilde{\alpha}}_i \right)\ ||\ \text{Dir}\left(\bpi | \langle1, ..., 1 \rangle \right) \bigr] ,
\end{align}
where $\boldsymbol{\tilde{\alpha}}_i = [\tilde{\alpha}_i^1, ..., \tilde{\alpha}_i^C]$ is the Dirichlet parameters after the removal of correct evidence from $\boldsymbol{\alpha}_i$ to penalize the misleading evidence only:
\begin{equation}
\begin{aligned}
    \tilde{\alpha}_i^c = 
        &\begin{cases}
            1 &\text{if } y_i^c = 1 \\
            \alpha_i^c &\text{if } y_i^c = 0 \\
        \end{cases} .
\end{aligned}
\end{equation}
The overall loss for our EDL segmentation network is
\begin{align}\label{EDL_OVERALL_LOSS}
    \calL = \calL_{\text{EDL}} +  \lambda_{\text{kl}} \calL_{\text{Reg}} , 
\end{align}
where $\lambda_{\text{kl}}$ is the hyperparameter.

Once the evidential model is trained to minimize the loss in \eqref{EDL_OVERALL_LOSS}, the belief mass $\bbi = [b_i^1, ..., b_i^C]$ and its uncertainty $u_i$ are reliably estimated for each $v_i$. By projecting these estimates into the point cloud, we obtain evidential points $\mathcal{D} = \{x_i, \mathbb{M}_i \}^N_{i=1}$, which are then utilized in the following mapping process, where $\mathbb{M}_i = \{ \bbi, u_i\}$ is the belief mass.

\subsection{Evidential Semantic Mapping}\label{METHOD_EVSemMap}
Unlike BKI-based mapping, which employs Bayes' rule to update the Dirichlet posterior of each map cell, our framework updates the semantic states via evidential fusion. Our mapping objective is to estimate belief mass $\mathbb{M}_* = \{ \bbstar, u_*\}$ for each position $x_*$ by using evidential points $\mathcal{D} = \{x_i, \mathbb{M}_i\}^N_{i=1}$. 

Based on DST~\cite{129_DST_yager2008classic}, we can integrate multiple belief masses by adopting the reduced Dempster's rule of combination~\cite{72_TMCJournal_han2022trusted}. The combination $\mathbb{M} = \{ \{ b^c \}_{c=1}^C, u \}$ of two belief masses $\mathbb{M}_1$ and $\mathbb{M}_2$, denoted by $\mathbb{M}_1 \oplus \mathbb{M}_2$, is computed as:
\begin{align}\label{dempster_fusion}
    b^c = \frac{1}{1-\delta} (b_1^c b_2^c + b_1^c u_2 + b_2^c u_1),\ \ u = \frac{1}{1-\delta} u_1 u_2,
\end{align}
where $\delta = \sum_{i \ne j} b_1^i b_2^j$ is a measure of the conflict between two belief masses. This fusion process enables direct consideration of semantic uncertainty from DNNs, leading to reliable quantification of map cell uncertainty.

However, the fusion process in~\eqref{dempster_fusion} fails to utilize local spatial information, which is essential for handling sparse sensor data, as seen in BKI-based mapping. To address this gap, we introduce an extended belief mass $\widetilde{\mathbb{M}}_i$, designed to broaden the impact of each evidential point from its original position $x_i$ to the surrounding space. This is achieved by applying a kernel function to the belief mass of the original evidential point as $k(x_i, x) * \bbi$, where the uncertainty $u_i$ is updated to satisfy the property in~\eqref{belief_sum}. Specifically, the adaptive sparse kernel is adopted based on \cite{kim2024evidential}:
\begin{equation}
\begin{aligned}\label{uBKI_kernel}
    k(x_i, x) = 
        \underset{u_{i} \le \Uthres}{\mathbbm{1}} k'(d, l \cdot \beta e ^{1 - \gamma u_i}, \sigma_0)
\end{aligned}
\end{equation}
where $\mathbbm{1}$ is the indicator function, $d$ indicates the distance between $x_i$ and $x$, $\Uthres$ is dynamically calculated to filter out top-$\psi$\% uncertain points, and $l$, $\sigma_0$, $\beta$, $\gamma$, $\psi$ are hyperparameters. The uncertainty threshold $\Uthres$ is dynamically calculated to filter out top-$\psi$\% uncertain points, and the sparse kernel $k'(d, l, \sigma_0)$~\cite{66_SparseKernel_melkumyan2009sparse} is defined as:
\begin{equation}\label{BKI_5_reparam}
\begin{aligned}
    k'(d, l, \sigma_0)=\underset{d < l}{\mathbbm{1}} \: \sigma_0 \Big[ \frac{2 + \cos (2\pi \dol)}{3} (1 - \dol) + \frac{1}{2\pi} \sin (2\pi \dol) \Big] .
\end{aligned}
\end{equation}

The extended belief mass $\widetilde{\mathbb{M}}_i^t$ at time $t$ can be integrated into the belief mass $\hat{\mathbb{M}}_*^t$ using the combination rule \eqref{dempster_fusion}:
\begin{align}\label{dempster_add}
    \hat{\mathbb{M}}^t_* = \widetilde{\mathbb{M}}^t_1 \oplus \widetilde{\mathbb{M}}^t_2 \oplus \cdots \oplus \widetilde{\mathbb{M}}^t_N.
\end{align}
Then, the semantic belief $\mathbb{M}^t_*$ can be recursively updated based on the semantic belief $\mathbb{M}^{t-1}_*$ at time $t - 1$ and the merged belief mass as $\mathbb{M}^t_*= \mathbb{M}^{t-1}_* \oplus \hat{\mathbb{M}}^{t}_*$. Based on the belief mass $\mathbb{M}^t_*$ of the evidential semantic map, the semantic state $s_* = \text{argmax}_c \bbstar$ and its uncertainty $u_*$ can be calculated to produce the final semantic map and its uncertainty map.

\section{EXPERIMENTS}
In this section, we validate that our method effectively enhances the reliability of semantic mapping and quantification of map cell uncertainty in perceptually challenging off-road settings using the same settings in our previous work~\cite{kim2024evidential}.

\subsection{Experimental Setup}
\noindent\textbf{Evaluation Metrics} 
Intersection over Union~($\mathrm{IoU}$), mean $\mathrm{IoU}$~($\mathrm{mIoU}$), and voxel semantic accuracy ($\mathrm{Acc}$) are employed to evaluate the semantic mapping performance. Additionally, Brier Score~($\mathrm{BS}$) is presented to evaluate the reliability of the uncertainty map~\cite{kim2024evidential}. This metric becomes lower if the uncertainty of correctly estimated cells is low and the uncertainty of incorrectly estimated map cells is high.

\noindent\textbf{Comparison Methods}
We perform comparative analyses with existing 3D continuous semantic mapping methods. \textit{S-BKI}~\cite{51_S-BKI_gan2020bayesian} is adopted as baselines for continuous semantic mapping. We also include results from advanced BKI-based mapping methods: \textit{ConvBKI}~\cite{47_ConvBKI2_wilson2023convbki}, \textit{SEE-CSOM}~\cite{88_SEE-CSOM_deng2023see}, and our previous work (\textit{\ourprevWork})~\cite{kim2024evidential}. While \textit{ConvBKI} considers class-specific geometry and \textit{SEE-CSOM} incorporates voxel semantic entropy, \textit{\ourprevWork} only introduces the DNN's semantic uncertainty into the BKI framework.

\begin{figure*}[t!]
\begin{center}
\includegraphics[width=1.0\linewidth]{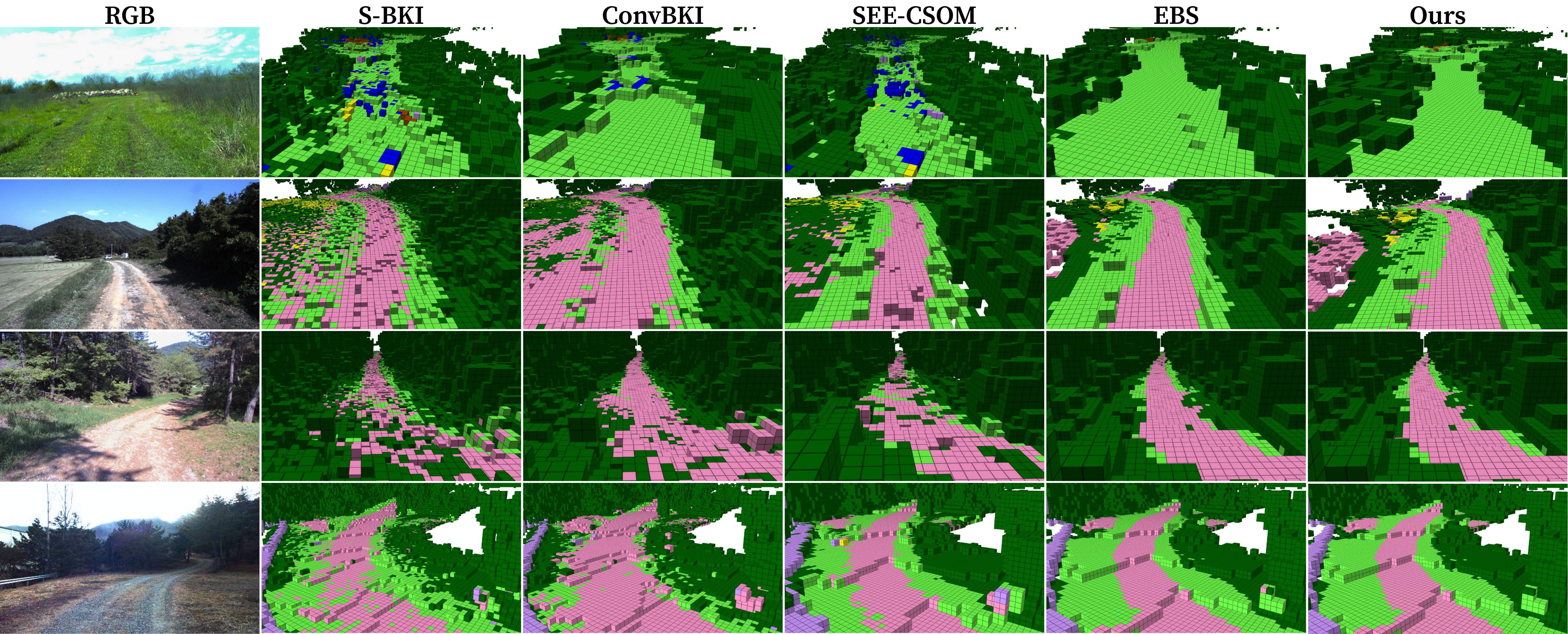}
\end{center}
\vspace{-0.1in}
\caption{
Qualitative results of 3D semantic mapping methods. Compared to others, our uncertainty-aware mapping methods (\textit{\ourprevWork}~\cite{kim2024evidential} and \textit{Ours}) generate accurate maps that preserve semantic details while excluding noisy predictions in both datasets. This is depicted by the plane of \semcolorbf{grass}{grass} in the first row and the boundaries of \semcolorbf{unpaved}{unpaved roads}, \semcolorbf{grass}{grass}, and \semcolorbf{vegetation}{vegetation} in the subsequent rows.
}
\label{fig:qualSEM}
\vspace{-0.1in}
\end{figure*}

\begin{table}[t!]
\centering
\renewcommand {\arraystretch}{1.15}
\caption{
Quantitative results on \textit{RELLIS-3D} and our \textit{OffRoad} dataset. Semantic classes not present in the evaluation dataset are excluded when calculating $\mathrm{mIoU}$ and denoted with a dash (-). Our method outperforms all 3D continuous mapping methods, excluding \textit{\ourprevWork}, with which it demonstrates comparable accuracy. Note that our method shows the lowest $\mathrm{BS}$ values, demonstrating the reliability of its map cell uncertainty. 
}
\label{tab:rellis_result}
\large{
\resizebox{1.0\linewidth}{!}{
    \begin{tabular}{c | c | ccccccc | cc |c}
    \toprule
    \textbf{Dataset}
    & \textbf{Method}
    & \rotatebox{90}{\semcolor[puddle]     \hspace{0pt}puddle}
    & \rotatebox{90}{\semcolor[object]     \hspace{0pt}object}
    & \rotatebox{90}{\semcolor[paved]      \hspace{0pt}paved}
    & \rotatebox{90}{\semcolor[unpaved]    \hspace{0pt}unpaved}
    & \rotatebox{90}{\semcolor[dirt]       \hspace{0pt}dirt}
    & \rotatebox{90}{\semcolor[grass]      \hspace{0pt}grass}
    & \rotatebox{90}{\semcolor[vegetation] \hspace{0pt}vegetation}
    & \rotatebox{90}{$\mathrm{mIoU}$ [\%]}
    & \rotatebox{90}{$\mathrm{Acc}$ [\%]} 
    & \rotatebox{90}{$\mathrm{BS}\ (\downarrow)$ [\%]}\\
    \midrule \midrule
\multirow{5}{*}{\shortstack{\textbf{\textit{RELLIS}} \\ \textbf{\textit{3D}}}}
    & \textit{S-BKI~\cite{51_S-BKI_gan2020bayesian}}&
    27.0 & 28.5 & \underline{53.8} & - & \textbf{13.4} & 76.4 & 73.8 & 45.5 & 83.7 & 35.7 \\
    
    & \textit{ConvBKI~\cite{47_ConvBKI2_wilson2023convbki}} &
    \textbf{30.6} & 22.4 & 45.5 & - & 6.3 & 69.6 & 68.4 & 40.5 & 79.7 & 35.5 \\
    
    & \textit{SEE-CSOM~\cite{88_SEE-CSOM_deng2023see}} &
    27.7 & 33.4 & 53.4 & - & 6.7 & 76.4 & \textbf{74.0} & 45.3 & 83.6 & 24.9 \\
    
    & \textit{\ourprevWork~\cite{kim2024evidential}} &
    \underline{29.7} & \underline{34.6} & \textbf{55.8} & - & \underline{12.1} & \textbf{76.7} & 73.8 & \textbf{47.1} & \textbf{84.0} & \underline{24.1} \\

    & \textit{Ours} & 
    26.3 & \textbf{36.8} & 52.8 & - & 8.0 & \underline{76.6} & \textbf{74.0} & \underline{45.8} & \underline{83.8} & \textbf{16.1} \\
    
    \hline
\multirow{5}{*}{\textit{\ourdatabf}} 
    & \textit{S-BKI~\cite{51_S-BKI_gan2020bayesian}}&
    - & 38.0 & - & 68.6 & - & 49.9 & 90.9 & 61.9 & 89.0 & 39.7 \\
    
    & \textit{ConvBKI~\cite{47_ConvBKI2_wilson2023convbki}} &
    - & 34.7 & - & 61.2 & - & 36.8 & 91.3 & 56.0 & 88.2 & 48.4 \\
    
    & \textit{SEE-CSOM~\cite{88_SEE-CSOM_deng2023see}} &
    - & 41.4 & - & 72.3 & - & 52.3 & 91.8 & 64.4 & 90.1 & 48.8 \\
    
    & \textit{\ourprevWork~\cite{kim2024evidential}} &
    - & \underline{44.2} & - & \textbf{76.4} & - & \textbf{57.9} & \textbf{92.5} & \textbf{67.8} & \textbf{91.3} & \underline{12.5} \\

    & \textit{Ours} &
    - & \textbf{45.0} & - & \underline{75.5} & - & \underline{55.9} & \textbf{92.5} & \underline{67.2} & \textbf{91.3} & \textbf{8.4} \\
    \bottomrule
    \end{tabular}
    }
}
\label{table:mainresult}
\vspace{-0.1in}
\end{table}

\subsection{Experimental Results}
\begin{figure*}[t!]
\begin{center}
\includegraphics[width=1.0\linewidth]{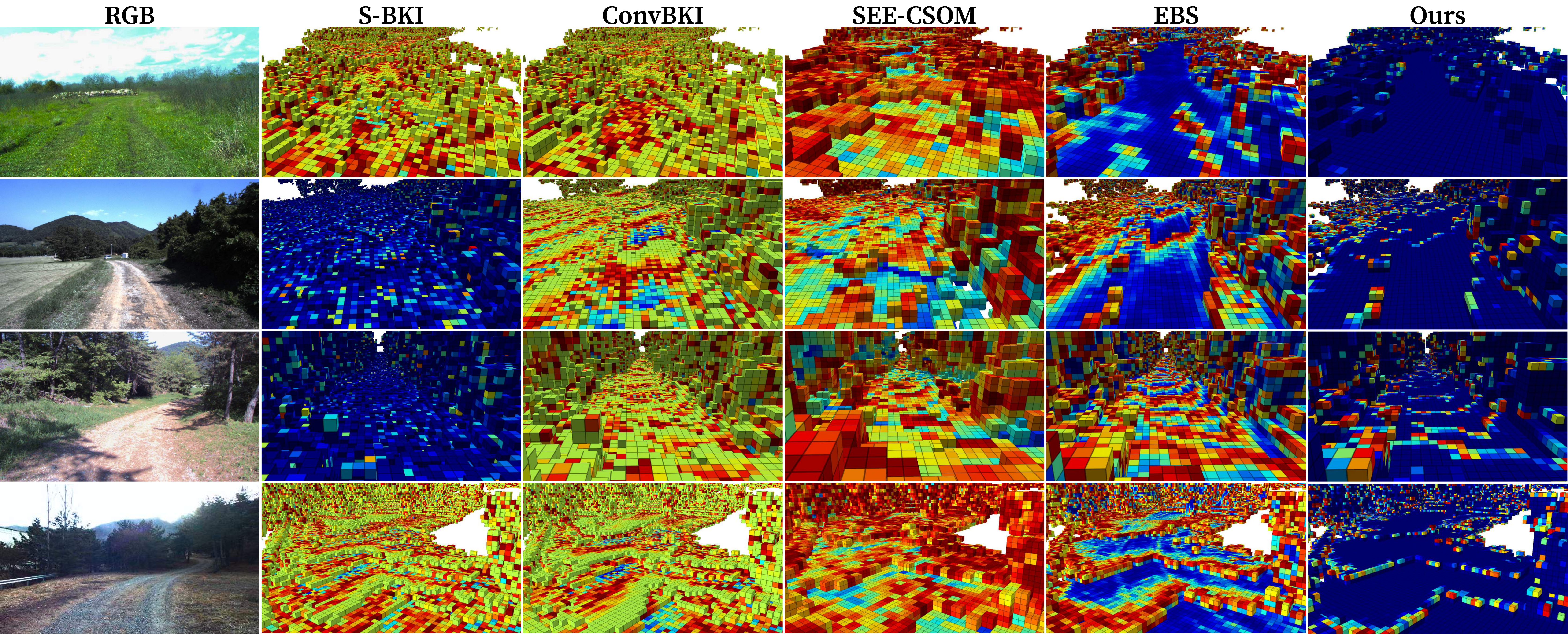}
\end{center}
\vspace{-0.1in}
\caption{
Visualization of uncertainty maps from 3D semantic mapping methods. The color of each cell represents the relative uncertainty value throughout each map. Though \textit{\ourprevWork} and ours only provide qualitatively meaningful uncertainty maps, \textit{\ourprevWork} tends to overestimate uncertainty. In contrast, our method effectively distinguishes between areas with high and low uncertainty.
}
\label{fig:qualVAR}
\vspace{-0.2in}
\end{figure*}

The quantitative results for \textit{RELLIS-3D} and our off-road dataset (\textit{OffRoad}) are presented in \Tref{table:mainresult}, where our framework demonstrates superior performance compared to all other continuous mapping techniques across both datasets, except for our previous work (\textit{\ourprevWork}), where it shows comparable performance. Furthermore, our method shows the lowest $\mathrm{BS}$ values among all 3D continuous mapping techniques, demonstrating the reliability of our map cell uncertainty.

\newcommand{\cmark}{\ding{51}}%
\begin{table}[b!]
\vspace{-0.2in}
\centering
\renewcommand{\arraystretch}{1.2}
\caption{
 Results of the ablation studies.
}
\label{tab:ablation}
\Large{
\resizebox{1.0\linewidth}{!}{%
    \begin{tabular}{cc cccc cccc}
        \toprule
        &&
        \multicolumn{4}{c}{\textbf{\textit{RELLIS-3D}}} & \multicolumn{4}{c}{\textit{\ourdatabf}} \\
        \cmidrule(lr){3-6} \cmidrule(lr){7-10} 
        \multirow{1}{*}{\textbf{Adap.}} & \multirow{1}{*}{\textbf{Thr.}} & $\mathrm{mIoU}$ & $\mathrm{Acc}$ & $\mathrm{O.Acc}$ & $\mathrm{BS}(\downarrow)$ & $ \mathrm{mIoU}$ & $\mathrm{Acc}$ & $\mathrm{O.Acc}$ & $\mathrm{BS}(\downarrow)$ \\
        \midrule \midrule
        - & \multicolumn{1}{c|}{-}& 
            \underline{46.4} & 83.3 & \underline{97.0} & 16.6 & 64.0 & 89.7 & \textbf{99.9} & 14.8 \\
        - & \multicolumn{1}{c|}{\cmark} & 
            \textbf{46.9} & 83.5 & 96.9 & 16.5 & \textbf{67.9} & \textbf{91.6} & 97.5 & 15.5 \\
        \cmark & \multicolumn{1}{c|}{-} & 
            45.0 & \underline{83.6} & \textbf{97.1} & \underline{16.3} & 63.3 & 90.0 & \textbf{99.9} & \underline{10.2} \\
        \cmark & \multicolumn{1}{c|}{\cmark} & 
            45.8 & \textbf{83.9} & \underline{97.0} & \textbf{16.1} & \underline{67.2} & \underline{91.3} & 97.9 & \textbf{8.4}  \\
        \bottomrule
    \end{tabular}
    }
}
\end{table}

\Fref{fig:qualSEM} and \Fref{fig:qualVAR} provide qualitative comparisons of 3D semantic maps and uncertainty maps generated by BKI-based continuous semantic mapping techniques and our evidential semantic mapping framework. The first row displays results from \textit{RELLIS-3D}, while the following rows show results from \textit{OffRoad}. While \textit{S-BKI}, \textit{ConvBKI}, and \textit{SEE-CSOM} exhibit more scattered distributions of noisy uncertainty values across cells, uncertainty-aware mapping frameworks, \textit{\ourprevWork} and ours, demonstrate more organized distinctions between areas of high and low uncertainty. However, \textit{\ourprevWork} tends to overestimate the uncertainty in the regions with correct semantics (\eg, planar regions with the same semantics) as it focuses on geometric uncertainty by indirectly considering semantic uncertainty based on BKI. In contrast, our uncertainty maps effectively differentiate between the regions with a low likelihood of correct semantics (\eg, boundaries between different semantic areas or distant regions from the sensors) and those with clear semantics.

Lastly, we conduct ablation studies to evaluate the impact of components in the adaptive kernel in~\eqref{uBKI_kernel}. Specifically, we examine the contribution of the adaptive kernel length (\textbf{Adap.}), and the uncertainty thresholding (\textbf{Thr.}) to the overall performance. The results are summarized in \Tref{tab:ablation}. In both datasets, the uncertainty threshold improves $\mathrm{mIoU}$, whereas the adaptive kernel length enhances the reliability of the uncertainty measure with lower $\mathrm{BS}$. Although the uncertainty threshold significantly improves the mapping performances, it exacerbates geometric completeness $\mathrm{O.Acc}$, defined as the proportion of queries with occupied voxels~\cite{kim2024evidential}. Combining the adaptive kernel length with the uncertainty threshold enhances semantic mapping performance and reliability while maintaining geometric completeness.

\section{CONCLUSIONS}
This work presents an evidential fusion framework for reliable semantic mapping in uncertain environments leveraging Dempster-Shafer Theory of Evidence. To address the challenges of high perceptual uncertainties, our mapping framework utilizes semantic uncertainty estimated by the EDL framework to accumulate semantic information for each map cell. Through extensive experiments, we demonstrate the effectiveness of our adaptive evidential fusion framework in handling semantic uncertainty, resulting in accurate semantic maps and reliable map cell uncertainty quantification in perceptually challenging unstructured outdoor environments.

\section*{ACKNOWLEDGMENT}
\normalsize{This work was supported by the Agency for Defense Development grant funded by Korean Government in 2024.}

\addtolength{\textheight}{0cm}   
                                  
\clearpage

\bibliographystyle{IEEEtran}
\bibliography{mybib.bib}

\begin{thebibliography}{10}
\providecommand{\url}[1]{#1}
\csname url@samestyle\endcsname
\providecommand{\newblock}{\relax}
\providecommand{\bibinfo}[2]{#2}
\providecommand{\BIBentrySTDinterwordspacing}{\spaceskip=0pt\relax}
\providecommand{\BIBentryALTinterwordstretchfactor}{4}
\providecommand{\BIBentryALTinterwordspacing}{\spaceskip=\fontdimen2\font plus
\BIBentryALTinterwordstretchfactor\fontdimen3\font minus \fontdimen4\font\relax}
\providecommand{\BIBforeignlanguage}[2]{{%
\expandafter\ifx\csname l@#1\endcsname\relax
\typeout{** WARNING: IEEEtran.bst: No hyphenation pattern has been}%
\typeout{** loaded for the language `#1'. Using the pattern for}%
\typeout{** the default language instead.}%
\else
\language=\csname l@#1\endcsname
\fi
#2}}
\providecommand{\BIBdecl}{\relax}
\BIBdecl

\bibitem{94_kim20133d}
B.-s. Kim, P.~Kohli, and S.~Savarese, ``3d scene understanding by voxel-crf,'' in \emph{IEEE/CVF International Conference on Computer Vision (CVPR)}, 2013, pp. 1425--1432.

\bibitem{103_valentin2013mesh}
J.~P. Valentin, S.~Sengupta, J.~Warrell, A.~Shahrokni, and P.~H. Torr, ``Mesh based semantic modelling for indoor and outdoor scenes,'' in \emph{IEEE/CVF Conference on Computer Vision and Pattern Recognition (CVPR)}, 2013, pp. 2067--2074.

\bibitem{95_SemanticOctree_sengupta2015semantic}
S.~Sengupta and P.~Sturgess, ``Semantic octree: Unifying recognition, reconstruction and representation via an octree constrained higher order mrf,'' in \emph{IEEE International Conference on Robotics and Automation (ICRA)}, 2015, pp. 1874--1879.

\bibitem{109_paz2020probabilistic}
D.~Paz, H.~Zhang, Q.~Li, H.~Xiang, and H.~I. Christensen, ``Probabilistic semantic mapping for urban autonomous driving applications,'' in \emph{IEEE/RSJ International Conference on Intelligent Robots and Systems (IROS)}, 2020, pp. 2059--2064.

\bibitem{23_SSMI_asgharivaskasi2023semantic}
A.~Asgharivaskasi and N.~Atanasov, ``Semantic octree mapping and shannon mutual information computation for robot exploration,'' \emph{IEEE Transactions on Robotics}, 2023.

\bibitem{26_morilla2023robust}
D.~Morilla-Cabello, L.~Mur-Labadia, R.~Martinez-Cantin, and E.~Montijano, ``Robust fusion for bayesian semantic mapping,'' in \emph{IEEE/RSJ International Conference on Intelligent Robots and Systems (IROS)}, 2023.

\bibitem{106_FusionOverconfidence_marques2023overconfidence}
J.~M.~C. Marques, A.~Zhai, S.~Wang, and K.~Hauser, ``On the overconfidence problem in semantic 3d mapping,'' \emph{IEEE International Conference on Robotics and Automation (ICRA)}, 2024.

\bibitem{50_GPOctoMap_wang2016fast}
J.~Wang and B.~Englot, ``Fast, accurate gaussian process occupancy maps via test-data octrees and nested bayesian fusion,'' in \emph{IEEE International Conference on Robotics and Automation (ICRA)}, 2016, pp. 1003--1010.

\bibitem{52_BKI_vega2014nonparametric}
W.~R. Vega-Brown, M.~Doniec, and N.~G. Roy, ``Nonparametric bayesian inference on multivariate exponential families,'' \emph{Advances in Neural Information Processing Systems (NeurIPS)}, vol.~27, 2014.

\bibitem{51_S-BKI_gan2020bayesian}
L.~Gan, R.~Zhang, J.~W. Grizzle, R.~M. Eustice, and M.~Ghaffari, ``Bayesian spatial kernel smoothing for scalable dense semantic mapping,'' \emph{IEEE Robotics and Automation Letters}, vol.~5, no.~2, pp. 790--797, 2020.

\bibitem{unnikrishnan2022dynamic}
A.~Unnikrishnan, J.~Wilson, L.~Gan, A.~Capodieci, P.~Jayakumar, K.~Barton, and M.~Ghaffari, ``Dynamic semantic occupancy mapping using 3d scene flow and closed-form bayesian inference,'' \emph{IEEE Access}, vol.~10, pp. 97\,954--97\,970, 2022.

\bibitem{wilson2023convolutional}
J.~Wilson, Y.~Fu, A.~Zhang, J.~Song, A.~Capodieci, P.~Jayakumar, K.~Barton, and M.~Ghaffari, ``Convolutional bayesian kernel inference for 3d semantic mapping,'' in \emph{IEEE International Conference on Robotics and Automation (ICRA)}, 2023, pp. 8364--8370.

\bibitem{96_SemanticFusion_mccormac2017semanticfusion}
J.~McCormac, A.~Handa, A.~Davison, and S.~Leutenegger, ``Semanticfusion: Dense 3d semantic mapping with convolutional neural networks,'' in \emph{IEEE International Conference on Robotics and automation (ICRA)}, 2017, pp. 4628--4635.

\bibitem{98_DA-RNN_xiang2017rnn}
Y.~Xiang and D.~Fox, ``Da-rnn: Semantic mapping with data associated recurrent neural networks,'' \emph{Robotics: Science and Systems (RSS)}, 2017.

\bibitem{99_RecurrentOctoMap_sun2018recurrent}
L.~Sun, Z.~Yan, A.~Zaganidis, C.~Zhao, and T.~Duckett, ``Recurrent-octomap: Learning state-based map refinement for long-term semantic mapping with 3-d-lidar data,'' \emph{IEEE Robotics and Automation Letters}, vol.~3, no.~4, pp. 3749--3756, 2018.

\bibitem{105_maturana2018real}
D.~Maturana, P.-W. Chou, M.~Uenoyama, and S.~Scherer, ``Real-time semantic mapping for autonomous off-road navigation,'' in \emph{Field and Service Robotics: Results of the 11th International Conference}, 2018, pp. 335--350.

\bibitem{wilson2022motionsc}
J.~Wilson, J.~Song, Y.~Fu, A.~Zhang, A.~Capodieci, P.~Jayakumar, K.~Barton, and M.~Ghaffari, ``Motionsc: Data set and network for real-time semantic mapping in dynamic environments,'' \emph{IEEE Robotics and Automation Letters}, vol.~7, no.~3, pp. 8439--8446, 2022.

\bibitem{86_OFFROAD_jin2021memory}
Y.~Jin, D.~Han, and H.~Ko, ``Memory-based semantic segmentation for off-road unstructured natural environments,'' in \emph{IEEE/RSJ International Conference on Intelligent Robots and Systems (IROS)}, 2021, pp. 24--31.

\bibitem{seo2023learning}
J.~Seo, S.~Sim, and I.~Shim, ``Learning off-road terrain traversability with self-supervisions only,'' \emph{IEEE Robotics and Automation Letters}, 2023.

\bibitem{kim2024evidential}
J.~Kim, J.~Seo, and J.~Min, ``Evidential semantic mapping in off-road environments with uncertainty-aware bayesian kernel inference,'' \emph{arXiv preprint arXiv:2403.14138}, 2024.

\bibitem{frey2024roadrunner}
J.~Frey, S.~Khattak, M.~Patel, D.~Atha, J.~Nubert, C.~Padgett, M.~Hutter, and P.~Spieler, ``Roadrunner--learning traversability estimation for autonomous off-road driving,'' \emph{arXiv preprint arXiv:2402.19341}, 2024.

\bibitem{107_guo2017calibration}
C.~Guo, G.~Pleiss, Y.~Sun, and K.~Q. Weinberger, ``On calibration of modern neural networks,'' in \emph{International Conference on Machine Learning (ICML)}, 2017, pp. 1321--1330.

\bibitem{83_ModernReliability_de2023reliability}
P.~de~Jorge, R.~Volpi, P.~H. Torr, and G.~Rogez, ``Reliability in semantic segmentation: Are we on the right track?'' in \emph{IEEE/CVF Conference on Computer Vision and Pattern Recognition (CVPR)}, 2023, pp. 7173--7182.

\bibitem{10_EDL_sensoy2018evidential}
M.~Sensoy, L.~Kaplan, and M.~Kandemir, ``Evidential deep learning to quantify classification uncertainty,'' \emph{Advances in Neural Information Processing Systems (NeurIPS)}, vol.~31, 2018.

\bibitem{47_ConvBKI2_wilson2023convbki}
J.~Wilson, Y.~Fu, J.~Friesen, P.~Ewen, A.~Capodieci, P.~Jayakumar, K.~Barton, and M.~Ghaffari, ``Convbki: Real-time probabilistic semantic mapping network with quantifiable uncertainty,'' \emph{arXiv preprint arXiv:2310.16020}, 2023.

\bibitem{Fan-RSS-21}
D.~D. Fan, K.~Otsu, Y.~Kubo, A.~Dixit, J.~Burdick, and A.~akbar Agha-mohammadi, ``{STEP: Stochastic Traversability Evaluation and Planning for Risk-Aware Off-road Navigation},'' in \emph{Proceedings of Robotics: Science and Systems (RSS)}, 2021.

\bibitem{rakowsky2007fundamentals}
U.~K. Rakowsky, ``Fundamentals of the dempster-shafer theory and its applications to reliability modeling,'' \emph{International Journal of Reliability, Quality and Safety Engineering}, vol.~14, no.~06, pp. 579--601, 2007.

\bibitem{122_RELLIS_jiang2021rellis}
P.~Jiang, P.~Osteen, M.~Wigness, and S.~Saripalli, ``Rellis-3d dataset: Data, benchmarks and analysis,'' in \emph{IEEE International Conference on Robotics and Automation (ICRA)}, 2021, pp. 1110--1116.

\bibitem{49_BGKOctoMap_doherty2017bayesian}
K.~Doherty, J.~Wang, and B.~Englot, ``Bayesian generalized kernel inference for occupancy map prediction,'' in \emph{IEEE International Conference on Robotics and Automation (ICRA)}, 2017, pp. 3118--3124.

\bibitem{88_SEE-CSOM_deng2023see}
Y.~Deng, M.~Wang, Y.~Yang, D.~Wang, and Y.~Yue, ``See-csom: Sharp-edged and efficient continuous semantic occupancy mapping for mobile robots,'' \emph{IEEE Transactions on Industrial Electronics}, 2023.

\bibitem{129_DST_yager2008classic}
R.~R. Yager and L.~Liu, \emph{Classic works of the Dempster-Shafer theory of belief functions}.\hskip 1em plus 0.5em minus 0.4em\relax Springer, 2008, vol. 219.

\bibitem{54_IEDL_deng2023uncertainty}
D.~Deng, G.~Chen, Y.~Yu, F.~Liu, and P.-A. Heng, ``Uncertainty estimation by fisher information-based evidential deep learning,'' \emph{International Conference on Machine Learning (ICML)}, 2023.

\bibitem{ulmer2023prior}
D.~T. Ulmer, C.~Hardmeier, and J.~Frellsen, ``Prior and posterior networks: A survey on evidential deep learning methods for uncertainty estimation,'' \emph{Transactions on Machine Learning Research}, 2023.

\bibitem{72_TMCJournal_han2022trusted}
Z.~Han, C.~Zhang, H.~Fu, and J.~T. Zhou, ``Trusted multi-view classification with dynamic evidential fusion,'' \emph{IEEE Transactions on Pattern Analysis and Machine Intelligence}, vol.~45, no.~2, pp. 2551--2566, 2022.

\bibitem{han2023ds}
J.~Han, Y.~Min, H.-J. Chae, B.-M. Jeong, and H.-L. Choi, ``Ds-k3dom: 3-d dynamic occupancy mapping with kernel inference and dempster-shafer evidential theory,'' in \emph{IEEE International Conference on Robotics and Automation (ICRA)}, 2023, pp. 6217--6223.

\bibitem{70_USNet_chang2022fast}
Y.~Chang, F.~Xue, F.~Sheng, W.~Liang, and A.~Ming, ``Fast road segmentation via uncertainty-aware symmetric network,'' in \emph{IEEE International Conference on Robotics and Automation (ICRA)}, 2022, pp. 11\,124--11\,130.

\bibitem{13_EvPSNet_sirohi2023uncertainty}
K.~Sirohi, S.~Marvi, D.~B{\"u}scher, and W.~Burgard, ``Uncertainty-aware panoptic segmentation,'' \emph{IEEE Robotics and Automation Letters}, vol.~8, no.~5, pp. 2629--2636, 2023.

\bibitem{120_EVORA_cai2023evora}
X.~Cai, S.~Ancha, L.~Sharma, P.~R. Osteen, B.~Bucher, S.~Phillips, J.~Wang, M.~Everett, N.~Roy, and J.~P. How, ``Evora: Deep evidential traversability learning for risk-aware off-road autonomy,'' \emph{arXiv preprint arXiv:2311.06234}, 2023.

\bibitem{130_SubjectiveLogic_jsang2018subjective}
A.~Jsang, \emph{Subjective Logic: A formalism for reasoning under uncertainty}.\hskip 1em plus 0.5em minus 0.4em\relax Springer Publishing Company, Incorporated, 2018.

\bibitem{66_SparseKernel_melkumyan2009sparse}
A.~Melkumyan and F.~T. Ramos, ``A sparse covariance function for exact gaussian process inference in large datasets,'' in \emph{International Joint Conference on Artificial Intelligence (IJCAI)}, 2009.

\end{thebibliography}


\end{document}